\documentclass[conference]{IEEEtran}
\IEEEoverridecommandlockouts
\usepackage{cite}
\usepackage{amsmath,amssymb,amsfonts}
\usepackage{algorithmic}
\usepackage{graphicx}
\usepackage{textcomp}
\usepackage{xcolor}
\usepackage{booktabs}
\usepackage{multirow}
\usepackage{tabularx}
\def\BibTeX{{\rm B\kern-.05em{\sc i\kern-.025em b}\kern-.08em
    T\kern-.1667em\lower.7ex\hbox{E}\kern-.125emX}}
\begin{document}

\title{Hyperspectral vs. RGB for Pedestrian Segmentation in Urban Driving Scenes: A Comparative Study\\
\thanks{This work has been submitted to the IEEE for possible publication. Copyright may be transferred without notice, after which this version may no longer be accessible.\\
\rule{0.75\columnwidth}{0.25pt}\\
~\textsuperscript{1} School of Engineering, University of Galway, Ireland \\
~\textsuperscript{2} Ryan Institute, University of Galway, Ireland \\
~\textsuperscript{3} Valeo Vision Systems, Tuam, Ireland\\
~\textsuperscript{\dag} Equal Contribution and correspondance (i.shah2@universityofgalway.ie)\\
}
}

\author{\IEEEauthorblockN{Jiarong Li~\textsuperscript{1,2,\dag}}
\and
\IEEEauthorblockN{Imad Ali Shah~\textsuperscript{1,2,\dag}}
\and
\IEEEauthorblockN{Enda Ward~\textsuperscript{3}}
\and
\IEEEauthorblockN{Martin Glavin~\textsuperscript{1,2}}
\and
\IEEEauthorblockN{Edward Jones~\textsuperscript{1,2}}
\and
\IEEEauthorblockN{Brian Deegan~\textsuperscript{1,2}}
}

\maketitle

\begin{abstract}
Pedestrian segmentation in automotive perception systems
faces critical safety challenges due to metamerism in RGB
imaging, where pedestrians and backgrounds appear visually
indistinguishable.. This study investigates the potential
of hyperspectral imaging (HSI) for enhanced pedestrian
segmentation in urban driving scenarios using the
Hyperspectral City v2 (H-City) dataset. We compared
standard RGB against two dimensionality-reduction
approaches by converting 128-channel HSI data into
three-channel representations: Principal Component Analysis
(PCA) and optimal band selection using Contrast
Signal-to-Noise Ratio with Joint Mutual Information
Maximization (CSNR-JMIM). Three semantic segmentation
models were evaluated: U-Net, DeepLabV3+, and SegFormer.
CSNR-JMIM consistently outperformed RGB with an average
improvements of 1.44\% in Intersection over Union (IoU) and
2.18\% in F1-score for pedestrian segmentation. Rider
segmentation showed similar gains with 1.43\% IoU and 2.25\%
F1-score improvements. These improved performance results
from enhanced spectral discrimination of optimally selected
HSI bands effectively reducing false positives. This study
demonstrates robust pedestrian segmentation through optimal
HSI band selection, showing significant potential for
safety-critical automotive applications.
\end{abstract}


\section{Introduction}
Ensuring the safety of pedestrians remains a critical objective in the development of Advanced Driver Assistance Systems (ADAS) and Autonomous Driving (AD)\cite{silva2025vulnerable}, with precise pedestrian segmentation increasingly recognized as a crucial factor for predicting movement patterns and safe navigation decisions in complex urban scenarios. However, conventional RGB-based segmentation systems face significant challenges due to metamerism phenomena~\cite{akbarinia2018color}, where pedestrians appear spectrally similar to the background (such as dark clothing against asphalt) under certain lighting conditions. This spectral ambiguity leads to segmentation failures at critical object boundaries, compromising the system's ability to accurately delineate pedestrians. Unlike detection, which requires identifying pedestrian presence, segmentation tasks demand precise object and material discrimination at the pixel level.

\begin{figure}[!t]
\centerline{\includegraphics[width=\columnwidth]{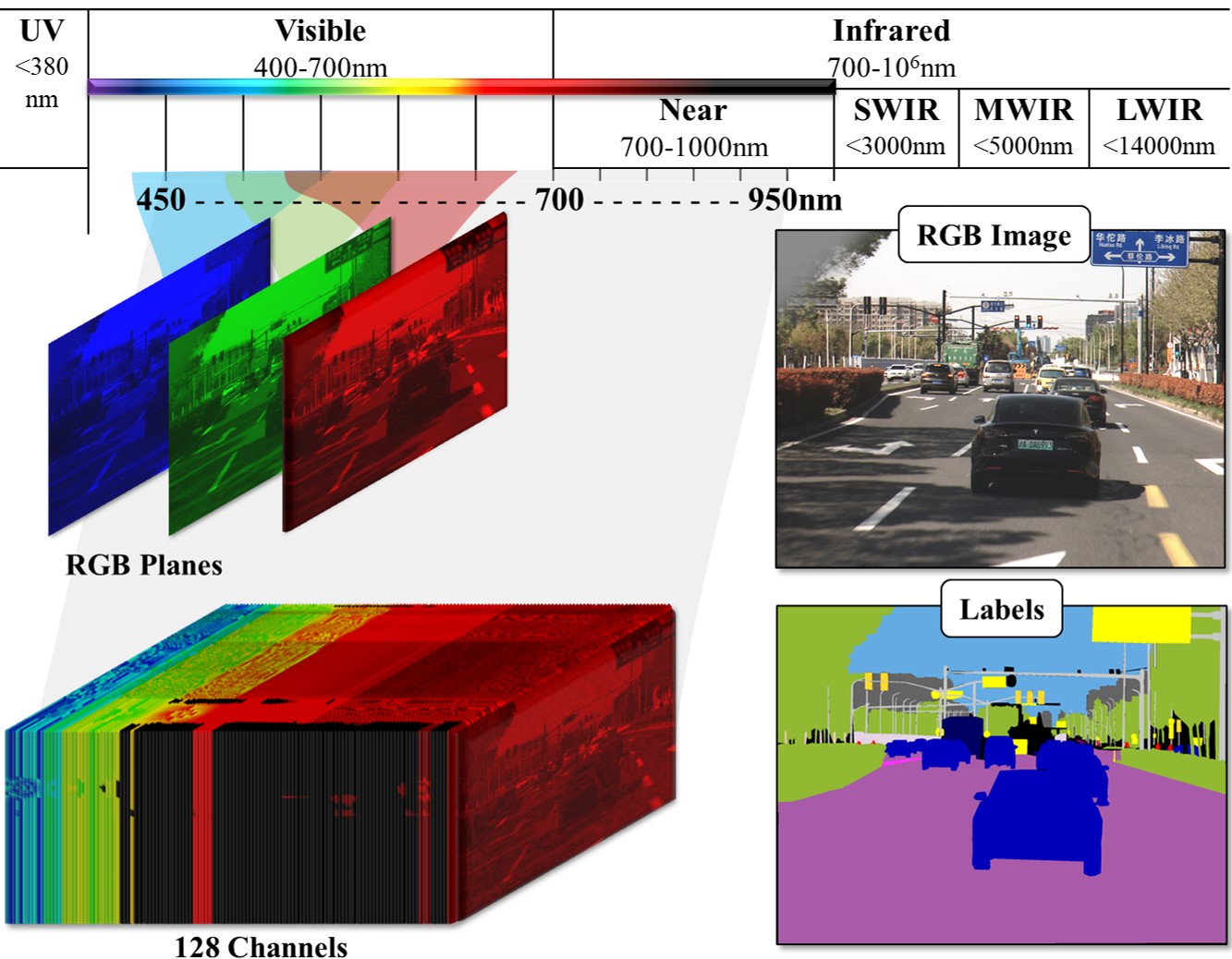}}
\caption{Comparison of RGB and HSI: Standard RGB captures three broad spectral bands, whereas HSI can acquire hundreds of narrow and contiguous spectral channels, providing a three-dimensional data, known as a hypercube. Sample RGB and hypercube are illustrated from the H-City dataset~\cite{shen4560035urban} with 128 channels, covering 450$\sim$950nm of the Electromagnetic Spectrum. Note: Color mapping is for representation purposes only.}
\label{fig1a:HSIvsRGB}
\end{figure}

Conventional perception systems typically rely on a single RGB sensor modality or a sensor suite of RGB, LiDAR, and radar. Although each sensor modality offers unique benefits, they struggle in scenarios with diverse lighting and weather conditions, sensors with their inherent limitations~\cite{van2018autonomous}, such as similar color-based ambiguity in RGB imaging. For instance, RGB can fail to distinguish a person with green clothing against vegetation or white clothing against a white wall in a certain lighting condition, even though their material properties are fundamentally different. LiDAR may show limitations in adverse weather conditions, and radar offers low-resolution data, making it difficult to differentiate a pedestrian from other objects~\cite{zhang2023perception}. This gap in perception capability requires the exploration of alternative sensing technologies that can provide a more robust representation of the scene. 

Hyperspectral Imaging (HSI) is one of the most promising candidates to overcome these challenges, such as metamerism and detection in adverse weather conditions~\cite{lu2020recent}. By capturing image data across hundreds of narrow, contiguous spectral bands, HSI provides a high-resolution spectral signature for each spatial pixel~\cite{lu2020recent}, as shown in Fig.~\ref{fig1a:HSIvsRGB}. HSI addresses these limitations by providing detailed information on the material composition of an object that is invisible to the human eye or RGB sensors~\cite{aburaed2023review}. Since pedestrian detection often depends on distinguishing clothing materials from background surfaces, their materials have unique spectral signatures that can be identified by an HSI sensor~\cite{driggers2013good}. This allows for segmentation based on intrinsic material properties rather than color information. However, the primary barrier to the widespread adoption of HSI in real-time applications is effective processing of its large amount of data (such as H-City: 128 bands vs 3 in RGB), which imposes significant computational demands~\cite{yuan2014hyperspectral}. One promising approach is to apply dimensionality-reduction techniques that preserve the most discriminative spectral information, which remains a key research objective for effective segmentation.

This paper presents a comparative study to determine whether dimensionality-reduced HSI data can offer a performance advantage over standard RGB images for pedestrian segmentation. The effectiveness of two dimensionality-reduction techniques is evaluated on H-City, one of the most spectrally diverse HSI datasets in ADAS/AD~\cite{shah2024hyperspectral}. The first approach is Principal Component Analysis (PCA)~\cite{abdi2010principal} and the second is an optimal band selection method~\cite{Li2025Toward} that combines Contrast
Signal-to-Noise Ratio (CSNR)~\cite{klein2023evaluation} and Joint Mutual Information Maximization (JMIM)~\cite{bennasar2015feature}. The resulting low-dimensional HSI datasets (pseudo-RGB), along with the corresponding RGB data, are then used to train and evaluate three well-established Semantic Segmentation Models (SSM): U-Net~\cite{ronneberger2015u}, DeepLabv3+~\cite{chen2018encoder}, and Segformer~\cite{xie2021segformer}. By comparing their performance, we provide a quantitative answer to whether the rich spectral information from HSI, even when compressed, provides a more robust foundation for pedestrian segmentation in comparison to standard RGB images. The main contributions of this work are:
\begin{itemize}
    \item Presenting the first comparative study that evaluates HSI versus standard RGB for pedestrian segmentation in ADAS/AD applications.
    \item Providing quantitative and qualitative evidence across three segmentation models (U-Net, DeepLabV3+, and SegFormer) that HSI can effectively address the metamerism challenge in pedestrian detection systems.
    \item Demonstrating that optimal band selection using CSNR-JMIM significantly outperforms both RGB and PCA-based dimensionality-reduction methods.
\end{itemize}

This paper is structured as follows: Section~\ref{Section:Related Work} provides related work in the fields of HSI in ADAS/AD contexts and the application of pedestrian segmentation. Section~\ref{Section:Methodology} explains the methodology for the input modalities and SSM. Section~\ref{Section:Experiments and Results} presents the experimental setup and a thorough analysis of the results. Section~\ref{Section:Conclusion} concludes with a discussion of our findings and their implications for the future of ADAS/AD perception systems.

\section{Related Work}
\label{Section:Related Work}
RGB-based pedestrian segmentation has been well researched, using traditional approaches as well as deep learning methods. Early works relied on handcrafted features such as the Histogram of Oriented Gradients (HOG)~\cite{dalal2005histograms}, while recent approaches leverage Convolutional Neural Networks (CNN) for enhanced feature extraction. Li et al. proposed a region-based CNN method that extends Faster R-CNN by incorporating SSM to reduce false positives in pedestrian detection~\cite{liu2018faster}. More recently, transformer-based architectures have shown promising results, with Vision Transformers (ViTs)~\cite{dosovitskiy2020image} demonstrating superior performance in complex scenarios. However, RGB-based methods still face significant challenges in varying and adverse lighting, shadows, and weather conditions, limiting their robustness in real-world applications~\cite{van2018autonomous,shah2024hyperspectral}.

To overcome the limitations of a single modality, researchers have explored multi-modal fusion. Techniques combining RGB with thermal (RGB-T) or depth (RGB-D) data have shown promising avenues by providing complementary information about the scene~\cite{peng2025tcainet, wu2024transformer}. Although beneficial, these methods still face challenges in complex scenarios and may not resolve ambiguities rooted in material properties. HSI has demonstrated potential use cases in the semantic understanding of visual information with applications in image segmentation, recognition, tracking, and pedestrian detection~\cite{zhang2022survey}, but its utilization in ADAS/AD remains limited~\cite{shah2025multi}. Researchers have demonstrated that the chemical properties of an object influence light frequency absorption and reflection~\cite{SheppardEnhanced2024}, making HSI's spectral information particularly valuable for material discrimination. The spectral reflectance of different objects in driving scenes beyond the visible spectrum can offer additional information to increase the reliability of the perception system, especially under challenging conditions~\cite{gutierrez2023chip}.

Recent HSI datasets for ADAS/AD perception, such as H-City~\cite{shen4560035urban}, aim to enhance research efforts for overall urban scenarios~\cite{shah2025multi}. However, while RGB methods currently dominate pedestrian detection systems, the comparative analysis of HSI and RGB for pedestrian segmentation remains unexplored. This gap motivates our study, which assesses the relative performance of RGB and HSI modalities for pedestrian segmentation in dynamic urban driving scenarios.

\begin{figure*}[htbp]
\centerline{\includegraphics[width=\textwidth]{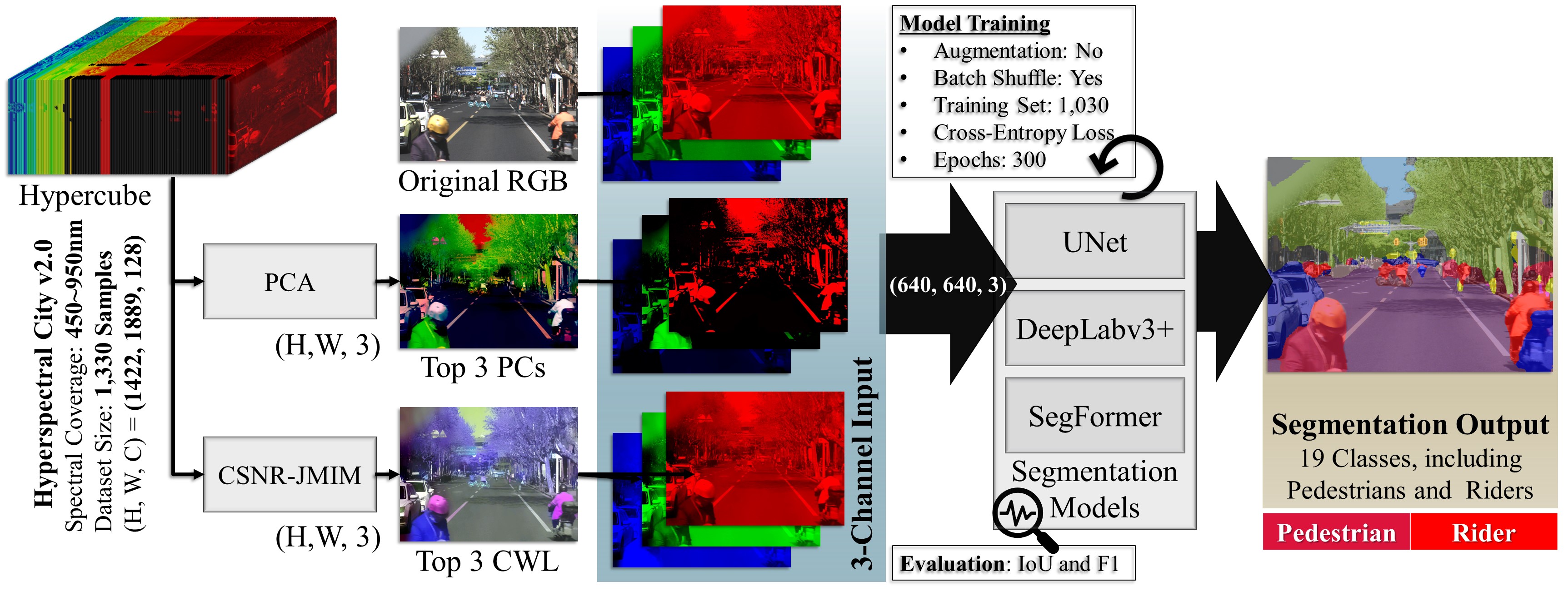}}
\caption{Comparative Pipeline for Pedestrian (including Riders) Segmentation Using RGB and HSI Inputs from the H-City Dataset. This workflow segments pedestrians from RGB and HSI data by processing hypercubes using dimensionality-reduction techniques, to three principal components through PCA and three Center Wavelengths (CWL) through CSNR-JMIM, to generate pseudo-RGB inputs. All input modalities are then trained on SSMs (UNet, DeepLabv3+, and SegFormer models) for evaluation of pedestrian segmentation accuracies.}
\label{fig1}
\end{figure*}

\section{Methodology}
\label{Section:Methodology}

Our methodology compares the performance of HSI and RGB data for pedestrian segmentation through a systematic evaluation framework. The core approach involves generating three-channel pseudo-RGB representations from the H-City dataset, which provides co-registered 128 spectral channels and standard RGB images of urban driving scenes. These pseudo-RGB datasets and standard RGB images are used and evaluated on three well-established SSM baseline models for direct performance comparison. It is to be noted that H-City is the only publicly available dataset that provides hypercubes co-registered with RGB images in an urban driving scene~\cite{shah2024hyperspectral}.

\subsection{Dimensionality Reduction based Pseudo-RGBs}\label{AA}

\textbf{Optimal Band Selection (CSNR-JMIM)}: We implement a multi-criteria band selection strategy that identifies the most informative spectral bands while minimizing redundancy~\cite{Li2025Toward}. This method combines Joint Mutual Information Maximization (JMIM)~\cite{bennasar2015feature}, inter-band correlation analysis, and Contrast Signal-to-Noise Ratio (CSNR)~\cite{klein2023evaluation} to select an optimal subset of three bands. For H-City, we selected wavelengths at 895 nm, 607 nm, and 497 nm, integrated over $\pm$27 nm, to generate the resulting pseudo-RGB image, which preserves maximum class-separable information while maintaining computational efficiency comparable to standard RGB processing.

\textbf{Principal Component Analysis (PCA)}~\cite{abdi2010principal}: As a baseline dimensionality-reduction method, PCA projects high-dimensional HSI data onto orthogonal components that maximize variance. To ensure consistency across the dataset, we calculated the top three principal components using a representative subset of 500 randomly sampled pixels from each of the 1,030 H-City training set. These three principal components constitute our second pseudo-RGB, i.e., transforming the H-City dataset into a 3-channel representation.

\subsection{Semantic Segmentation Models (SSM)}

To ensure robust and unbiased evaluation, we assess all input modalities using three established SSMs representing both CNN and Transformer-based architectures:

\textbf{U-Net}~\cite{ronneberger2015u} is a CNN-based, symmetric encoder-decoder structure that effectively preserves fine-grained spatial details, resulting in precise boundary segmentation. Its efficiency and strong performance with limited training data make it an excellent baseline for comparison. We implemented a 4-stage U-Net with encoder-decoder channels of 64, 128, 512, and 1024, connected via skip connections.

\textbf{DeepLabV3+}~\cite{chen2018encoder} is also a CNN-based model with atrous spatial pyramid pooling to capture multi-scale contextual information. The decoder module upsamples features, concatenates with low-level representations, and refines segmentation boundaries for improved accuracy. Similar to U-Net, Deeplabv3+ is also a widely adopted architecture for general-purpose semantic segmentation.

\textbf{SegFormer}~\cite{xie2021segformer} represents a Transformer-based SSM, combining a hierarchical transformer encoder with a lightweight multi-layer perceptron decoder. This architecture efficiently generates multi-scale features without requiring complex positional encodings, achieving an optimal balance between performance and efficiency.

The methodology consisted of nine distinct training and evaluation pipelines (three input modalities and three base SSMs), with SegFormer variants of Mix Transformer (MiT) b0, b3, and b5, providing extended comparison. This resulted in 15 total model training and evaluation configurations. Each model was trained from scratch using identical hyperparameters to ensure fair comparison. Performance evaluation employed Intersection over Union (IoU), F1-scores, Precision (Prec), and Recall (Rec) metrics, with particular focus on pedestrian and rider classes critical for ADAS/AD safety applications.

\begin{table*}[h]
\caption{Quantitative Analysis of input modalities (RGB, and pseudo-RGBs of PCA and CSNR-JMIM) across SSMs based on IoU, F1, Precision (Prec), and Recall (Rec)}
\centering
\renewcommand{\arraystretch}{1.2}
\scriptsize
\begin{tabular*}{\textwidth}{@{\extracolsep{\fill}}ll|cccc|cccc|cccc}
\hline
\textbf{Models} & \textbf{Input Modality} & \multicolumn{4}{c|}{\textbf{Overall 19 Classes}} & \multicolumn{4}{c|}{\textbf{Pedestrian Class}} & \multicolumn{4}{c}{\textbf{Rider Class}} \\
 & & \textbf{mIoU} & \textbf{mF1} & \textbf{mPrec} & \textbf{mRec} & \textbf{IoU} & \textbf{F1} & \textbf{Prec} & \textbf{Rec} & \textbf{IoU} & \textbf{F1} & \textbf{Prec} & \textbf{Rec} \\
\hline
U-Net~\cite{ronneberger2015u} & RGB & \textbf{47.09} & \textbf{58.36} & \textbf{59.57} & \textbf{59.81} & 18.93 & 31.84 & \textbf{27.43} & 37.94 & \textbf{17.08} & \textbf{29.18} & 29.86 & \textbf{28.54} \\
 & PCA~\cite{abdi2010principal} & 34.16 & 44.80 & 49.26 & 43.80 & 9.65 & 17.60 & 13.01 & 27.19 & 7.67 & 14.25 & 16.87 & 12.34 \\
 & CSNR-JMIM~\cite{Li2025Toward} & 44.27 & 55.57 & 57.52 & 56.95 & \textbf{20.40} & \textbf{33.88} & 25.48 & \textbf{50.57} & 16.74 & 28.68 & \textbf{32.62} & 25.59 \\
\hline
DeepLabv3+~\cite{chen2018encoder} & RGB & \textbf{47.18} & \textbf{58.88} & \textbf{60.85} & \textbf{59.84} & 21.51 & 35.41 & 30.06 & \textbf{43.07} & 20.70 & 34.30 & 35.33 & 33.33 \\
 & PCA & 38.84 & 50.40 & 53.72 & 50.20 & 14.22 & 24.90 & 21.18 & 30.20 & 11.94 & 21.34 & 24.93 & 18.65 \\
 & CSNR-JMIM & 45.82 & 57.71 & 59.24 & 58.97 & \textbf{21.99} & \textbf{36.05} & \textbf{32.02} & 41.25 & \textbf{22.47} & \textbf{36.69} & \textbf{36.75} & \textbf{36.63} \\
\hline
Segformer~\cite{xie2021segformer} & RGB & 35.65 & 46.27 & 48.64 & \textbf{47.42} & 9.84 & 17.91 & 12.94 & \textbf{29.10} & 7.25 & 13.52 & 13.96 & 13.10 \\
(MiT-b0) & PCA & 28.57 & 38.22 & 41.66 & 37.76 & 5.10 & 9.70 & 8.56 & 11.18 & 7.39 & 13.77 & 13.94 & 13.60 \\
 & CSNR-JMIM & \textbf{35.92} & \textbf{46.71} & \textbf{49.69} & 47.08 & \textbf{11.00} & \textbf{19.81} & \textbf{15.96} & 26.12 & \textbf{10.90} & \textbf{19.65} & \textbf{20.19} & \textbf{19.14} \\
\hline
Segformer & RGB & \textbf{39.72} & \textbf{50.80} & 52.52 & \textbf{52.03} & 11.67 & 20.91 & 20.29 & 21.56 & 13.52 & 23.83 & 20.32 & \textbf{28.80} \\
(MiT-b3) & PCA & 31.58 & 41.87 & 46.24 & 40.64 & 9.33 & 17.06 & 13.89 & 22.12 & 6.24 & 11.74 & 13.04 & 10.67 \\
 & CSNR-JMIM & 39.62 & 50.67 & \textbf{52.94} & 51.20 & \textbf{14.13} & \textbf{24.76} & \textbf{21.09} & \textbf{29.96} & \textbf{14.51} & \textbf{25.35} & \textbf{27.31} & 23.65 \\
\hline
Segformer & RGB & \textbf{41.31} & \textbf{52.57} & \textbf{54.92} & \textbf{53.24} & 14.68 & 25.60 & 21.16 & 32.39 & 13.67 & 24.05 & 21.82 & 26.78 \\
(MiT-b5) & PCA & 32.42 & 42.74 & 46.80 & 41.63 & 8.95 & 16.42 & 13.58 & 20.77 & 8.61 & 15.86 & 15.80 & 15.92 \\
 & CSNR-JMIM & 39.69 & 50.75 & 53.30 & 50.92 & \textbf{16.33} & \textbf{28.07} & \textbf{23.54} & \textbf{34.78} & \textbf{14.77} & \textbf{25.73} & \textbf{24.53} & \textbf{27.06} \\
\hline
\end{tabular*}
\label{table:results}
\end{table*}

\begin{figure*}[!ht]
\centerline{\includegraphics[width=\textwidth]{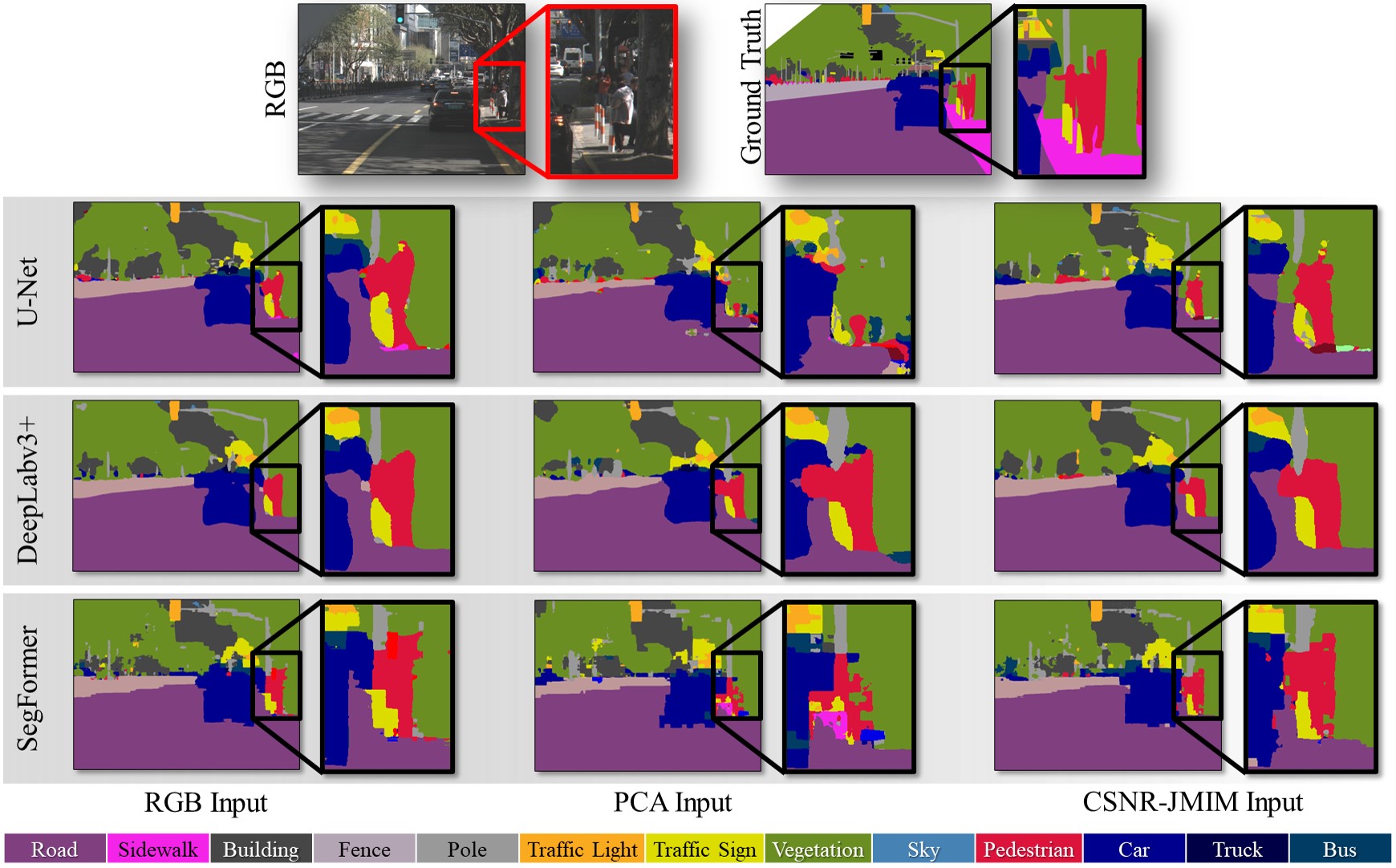}}
\caption{Segmentation outputs for all modalities and SSMs on a representative image from the H-City dataset with varying light conditions. The top row presents the original RGB image and the corresponding ground truth mask. Subsequent rows display segmentations from U-Net, DeepLabV3+, and SegFormer models, respectively. For each model, results are shown across three input modalities: RGB, PCA-based pseudo-RGB, and CSNR-JMIM-based pseudo-RGB.}
\label{fig2}
\end{figure*}

    

\section{Experiments and Results}
\label{Section:Experiments and Results}

Experiments were conducted using 1,330 hypercubes from the H-City dataset, following the original train-test split of 1,030 training samples. Hypercubes were preprocessed using PCA and CSNR-JMIM to generate three-channel representations for comparison with standard RGB.

All models were trained on dual NVIDIA RTX A6000 GPUs with 256GB RAM and Intel Xeon Gold 6252 processors. Training parameters included 300 epochs with AdamW optimizer, 0.0001 learning rate, and 16 batch size. To ensure a fair comparison across input modalities, no data augmentation, regularization, or preprocessing techniques were applied, other than resizing the input images to 640\texttimes640 pixels to balance computational efficiency.

\begin{figure}[t!]
\centerline{\includegraphics[width=0.498\textwidth]{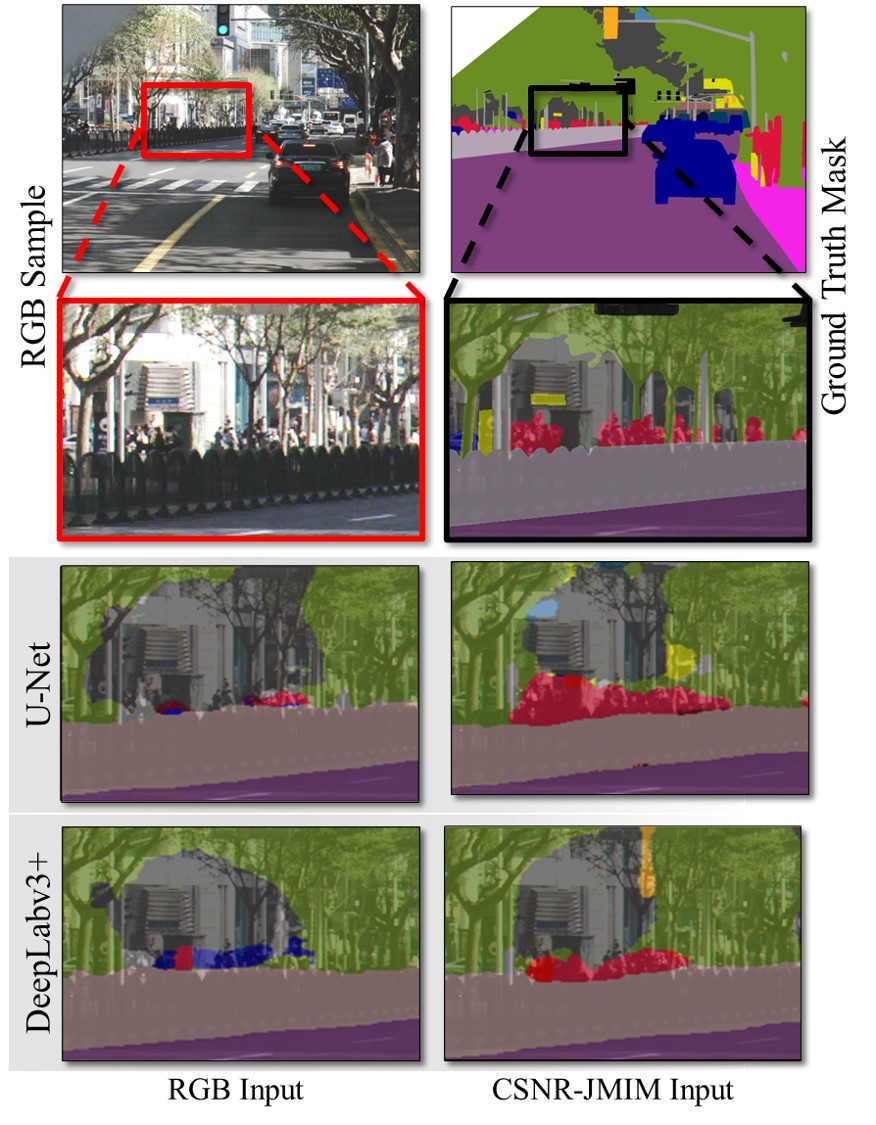}}
\caption{Comparison of semantic segmentation performance for small objects using RGB versus CSNR-JMIM input modalities. Row 1 shows the sample RGB image with its ground truth mask. Row 2 provides a magnified view focusing on pedestrians under challenging and variable lighting conditions. Rows 3 and 4 present segmentation outputs from U-Net and DeepLabv3+ models, respectively.}
\label{fig3.UNetVsDeepLab}
\end{figure}
\begin{figure}[t!]
\centerline{\includegraphics[width=0.498\textwidth]{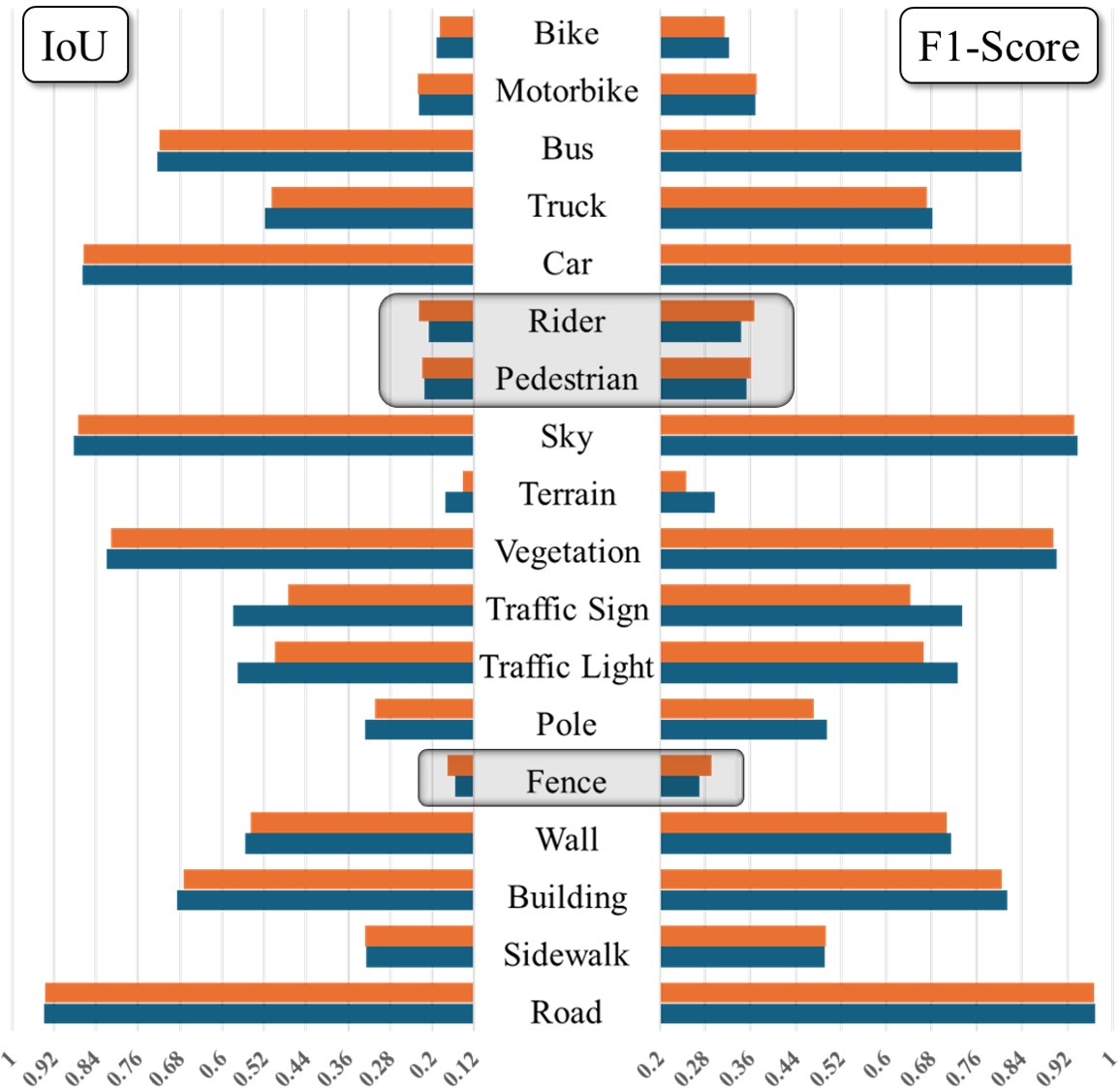}}
\caption{Classwise IoU and F1-Scores comparison of DeepLabv3+ with RGB (blue) and CSNR-JMIM (orange) Inputs.}
\label{fig4.ClassWiseComparison}
\end{figure}

\subsection{Quantitative Performance Evaluation}
The quantitative results in Table~\ref{table:results} highlight a key trade-off between overall semantic segmentation performance and the accurate detection of safety-critical classes. While RGB input generally achieves the highest mean performance across all 19 semantic classes in the H-City dataset, it consistently underperforms in segmenting pedestrian and rider classes, which are among the most crucial categories for AD safety.

CSNR-JMIM consistently improves performance on pedestrians and riders across SSMs, including both CNN-based (U-Net, DeepLabV3+) and Transformer-based (SegFormer) models. Specifically, CSNR-JMIM provides average improvements of 1.44 \% IoU and 2.22 \% F1 across SSMs, with improvements of 1.44 \% IoU/2.18 \% F1 for pedestrians and 1.43 \% IoU/2.25 \% F1 for riders. These improvements are a result of CSNR-JMIM’s ability to retain spectrally discriminative information that better distinguishes between visually similar regions, effectively mitigating the limitations of RGB-based systems caused by spectral metamerism.

In comparison, PCA underperforms consistently across all models and metrics. This suggests that variance-based dimensionality-reduction struggles to preserve the task-relevant spectral information necessary for fine-grained semantic segmentation.

The consistent performance improvement with CSNR-JMIM for pedestrians and riders, across SSMs and metrics, demonstrates its robustness and generalizability. This demonstration makes CSNR-JMIM valuable in safety-critical applications, where even subtle gains in pedestrian and rider segmentation can significantly impact real-world ADAS/AD effectiveness.

\subsection{Qualitative Analysis}
Fig.~\ref{fig2}--\ref{fig3.UNetVsDeepLab} visually confirms the quantitative findings, demonstrating that PCA consistently underperforms against RGB and CSNR-JMIM input modalities across all SSMs. The qualitative results show that CSNR-JMIM improves pedestrian segmentation with enhanced boundary delineation and reduced false positives:
\begin{itemize}
    \item \textbf{DeepLabv3+} shows the most significant improvement with CSNR-JMIM, achieving substantially reduced false positives. UNet and SegFormer also demonstrate improvements in pedestrian boundary delineation, though with less precision than DeepLabv3+.
    \item \textbf{CSNR-JMIM} enhances boundary precision, as shown in Fig.~\ref{fig3.UNetVsDeepLab}, reduces background misclassification, and improves detection of small or partially occluded pedestrians relative to RGB input.
    \item In addition to pedestrian and riders, CSNR-JMIM also provides better fence segmentation performance, as shown in the class-wise comparison of DeepLabv3+ in Fig.~\ref{fig4.ClassWiseComparison}. However, Traffic light and sign classes show degraded performance, indicating the need for future studies to incorporate all class-based band selection strategies.
\end{itemize}
The enhanced spectral discrimination of CSNR-JMIM enables better material-based distinction between pedestrians and background objects, effectively addressing lighting variations and metamerism challenges in scenarios where clothing closely matches background surfaces, as demonstrated in Fig.~\ref{fig3.UNetVsDeepLab}. These qualitative results support the quantitative analysis, demonstrating HSI's clear potential for enhanced pedestrian segmentation in ADAS/AD applications.

\section{Conclusion}
\label{Section:Conclusion}
This study presents the first comparison of HSI versus RGB modalities for pedestrian segmentation in the ADAS/AD domain. Our experimental evaluation across three SSMs demonstrates that HSI data significantly outperforms standard RGB imaging for pedestrian and rider segmentation. The CSNR-JMIM band selection approach consistently achieves improved performance compared to RGB across all evaluated models, with IoU improvements of 1.44 \% and 1.43 \%, and F1 score improvements of 2.18 \% and 2.25 \% for pedestrian and rider classes, respectively. These improvements result from HSI's ability to capture intrinsic material properties that remain discriminative across varying lighting conditions, effectively addressing the metamerism challenge that limits RGB-based systems. Moreover, the underperforming PCA compared to both RGB and CSNR-JMIM demonstrates that careful preservation of discriminative spectral information is critical for effective dimensionality-reduction in HSI processing.

The contributions of this work extend beyond pedestrian segmentation to broader ADAS/AD perception systems, where optimal spectral band selection can enhance overall segmentation performance across multiple object classes. As HSI technology advances and becomes more accessible for automotive applications, our findings indicate significant potential for improving road safety through more robust perception capabilities. Future research should focus on developing real-time implementation strategies and evaluating them under diverse weather conditions, including fog and haze.

This research establishes a foundation for transitioning from RGB-dominated perception systems to more robust HSI-based approaches, demonstrating how spectral imaging can enhance safety-critical applications where accurate pedestrian detection remains essential.


\bibliographystyle{IEEEtran}
\bibliography{references}
\end{document}